[Title page]

# Product versus Process: Exploring EFL Students' Editing of AI-generated Text for Expository Writing

***David James Woo*** (Precious Blood Secondary School, Hong Kong, China)

***Yangyang Yu*** (Shanghai Jiao Tong University, Shanghai, China)

***Kai Guo*** (The University of Hong Kong, Hong Kong, China)

***Yilin Huang*** (The Education University of Hong Kong, Hong Kong, China)

***April Ka Yeng Fung*** (Hoi Ping Chamber of Commerce Secondary School, Hong Kong, China)

**Corresponding author**
- Name: Yangyang Yu
- Email address: florayu0209@sjtu.edu.cn
- Postal address: School of Foreign Languages, Shanghai Jiao Tong University, No. 800 Dongchuan Road, Minhang District, Shanghai, China, 200240

**Declarations of interests**

The authors report there are no competing interests to declare.

**Data availability statement**

The data supporting this study's findings are available from the first author, David James Woo, upon reasonable request.

**Acknowledgements**

The authors express their gratitude to Dr. Chi Ho Yeung and Dr. Hengky Susanto for their indirect support of the study's data preparation and analysis.

# Product versus Process: Exploring EFL Students' Editing of AI-generated Text for Expository Writing

**Abstract**

**Background:** Text generated by artificial intelligence (AI) chatbots is increasingly used in English as a foreign language (EFL) writing contexts, yet its impact on students' expository writing process and compositions remains understudied.

**Objectives:** This research examines how EFL secondary students edit AI-generated text, exploring editing behaviors in their expository writing process and in expository compositions, and their effect on human-rated scores for content, organization, language, and overall quality.

**Methods:** Participants were 39 Hong Kong secondary students who wrote an expository composition with AI chatbots in a workshop. A convergent design was employed to analyze their screen recordings and compositions to examine students' editing behaviors and writing qualities. Analytical methods included qualitative coding, descriptive statistics, temporal sequence analysis, human-rated scoring and multiple linear regression (MLR) analysis.

**Results and Conclusions:** We analyzed over 260 edits per dataset, and identified two editing patterns: one where students refined introductory units repeatedly before progressing, and another where they quickly shifted to extensive edits in body units (e.g., topic and supporting sentences). MLR analyses revealed that the number of AI-generated words positively predicted all score dimensions, while most editing variables showed minimal impact. These results suggest a disconnect between students' significant editing effort and improved composition quality, indicating AI supports but does not replace writing skills.

**Implications:** The findings highlight the importance of genre-specific instruction and process-focused writing before AI integration. Educators should also develop assessments valuing both process and product to encourage critical engagement with AI text.

**Lay summary**

**What is currently known about this topic?**

- ✓ Expository writing is a crucial but challenging skill for EFL students.
- ✓ Generative AI tools such as ChatGPT can help students write compositions and improve writing performance.

- Few studies have been conducted to understand how students use AI-generated text during the writing process.

**What does this paper add?**

- It sheds lights on students' considerable editing effort while integrating AI-generated text into their expository compositions,
- It points to students' limited improvement in their final expository compositions even with the help of AI-generated text.

**Implications for practice/or policy**

- EFL educators should teach students foundational writing knowledge such as expository writing principles.
- EFL educators should guide students to enhance their writing with generative AI and reflect on how to use AI-generated text wisely.

## 1. Introduction

Expository writing provides information, explains topics, or clarifies concepts in a clear and straightforward manner (Slater & Graves, 1989). It is commonly found in textbooks, instruction manuals, essays, articles, and other informative texts. Although expository writing is a crucial skill for language learners if they are to progress academically and to communicate information effectively (Nippold, 2016; Roohani & Taheri, 2015), mastering it can be challenging for these students (Meisuo, 2000), due to factors such as limited language proficiency, underdeveloped writing strategies, and insufficient exposure to the genre (Hidi et al., 2006; Sasaki & Hirose, 1996; Yang et al., 2023).

Generative artificial intelligence (AI) chatbots (e.g., ChatGPT, Claude, Gemini, and Llama) can generate extensive information to enhance students' writing processes (Yang et al., 2022; Godwin-Jones, 2024). Specifically, students may improve the organization, coherence, grammar, and vocabulary of their written compositions (Song & Song, 2023; Tsai et al., 2024) with a 'machine-in-the-loop', whereby students collaborate with AI tools that provide timely assistance in the planning, drafting, and revising stages of process writing (Guo & Li, 2024; Su et al., 2023). However, this innovative method also poses potential risks to writing development, such as concerns about learning loss, diminished authorial voice, and academic integrity (Niloy et al., 2024; Zou & Huang, 2024). Some educators may find this innovative writing method threatening when they cannot identify the extent to which students have pasted AI-generated text into their compositions (Alexander et al., 2023). Additionally, few students have shown awareness that overreliance on AI-generated text in compositions can be unethical and may undermine their writing skills (Šedlbauer et al., 2024; Wang, 2024).

Investigating students' cognitive activities (Bennett et al., 2020; Vandermeulen et al., 2024) during machine-in-the-loop writing and their impact on completed compositions remains largely unexplored. To build knowledge on machine-in-the-loop writing in language learning contexts, we explore English as a foreign language (EFL) students' editing behaviors when incorporating AI-generated text into their expository writing compositions. We also examine these behaviors' influence on composition quality. Evidence of how EFL students are writing with a machine-in-the-loop and the subsequent effects on their compositions can inform educators' expectations and strategies for ethical and effective integration of generative AI tools in expository writing pedagogy.

## 2. Literature Review

### 2.1. Expository Writing

Expository writing is a fundamental means by which language learners exhibit the knowledge they have acquired (Nippold, 2016). Slater and Graves (1989) define expository text as a form of writing that is informative, explanatory, and straightforward. They assert that this type of text actively involves readers by emphasizing key information for their understanding. In contrast to other genres, such as narratives, expository writing demands more sophisticated linguistic tools from language learners (Lundine & McCauley, 2016). Additionally, as EFL learners advance academically, expository writing becomes increasingly significant (Roohani & Taheri, 2015). For instance, proficiency in expository writing can impact learning in other subjects like mathematics and computational thinking (Craig, 2016; Santos & Semana, 2015; Wolz et al., 2011). However, research indicates a prevalent use of narrative text over expository text in early education (Duke, 2000). This trend can lead to students having less familiarity with the expository genre (Ness, 2011). As a result, many children might exhibit a lower proficiency in writing expository text compared to other types of writing. Additionally, studies have highlighted students' struggles in producing high-quality expository essays and the necessity of training them in expository writing. For instance, Meisuo (2000) investigated the use of cohesive features in Chinese EFL students' expository compositions. The study highlighted specific issues, such as reference ambiguity, excessive or incorrect use of conjunctions, and limited lexical cohesion. A possible explanation for students' expository writing struggles is that students are not taught to write expository essays. Instead, students are taught to write narrowly defined text types that can include expository writing and other types of writing (Koh, 2015; The Curriculum Development Council, 2017). Regrettably, while extensive research exists on teaching children to *read* expository texts, there is comparatively less information on the best practices for teaching children to *write* expository texts (Fang, 2014).

Notably, generative AI technologies have been increasingly utilized to support argumentative writing (Guo & Li, 2024; Su et al., 2023; Zhang et al., 2024), where AI can support developing claims and evidence structures, and narrative writing (Woo et al., 2024a; Woo et al., 2024b), where AI can support storytelling elements. However, research remains limited on leveraging AI to support expository writing, where AI could support providing accurate factual information and clear explanatory structures, for instance. The unique demands of expository writing may create differences in how students use and edit AI-generated text compared to how they edit persuasive or creative content. Further exploration in this area is warranted to fully understand the potential of AI technologies in improving expository writing skills.

## *2.2. Machine-in-the-loop Writing*

Generative AI enables a 'machine-in-the-loop' approach to writing, whereby a student collaborates with a generative AI tool to compose a written work (Clark et al., 2018; Yang et al., 2022). As depicted in Figure 1, this cyclical process begins with a student providing instructions or prompts, for instance, a question, a command, or an excerpt of text, to guide the generative AI tool. The AI then generates text-based output based on its interpretation of the student's prompt. The student evaluates this output and decides whether to insert all, some or none of it in their written work. Independently, the student is able to complete the written work, composing text without AI assistance, or modifying existing AI-generated text in the work. The cycle can occur in the planning, drafting and revising phases of process writing. The cycle can repeat until the written work is completed, with the student retaining full control over the completed work.

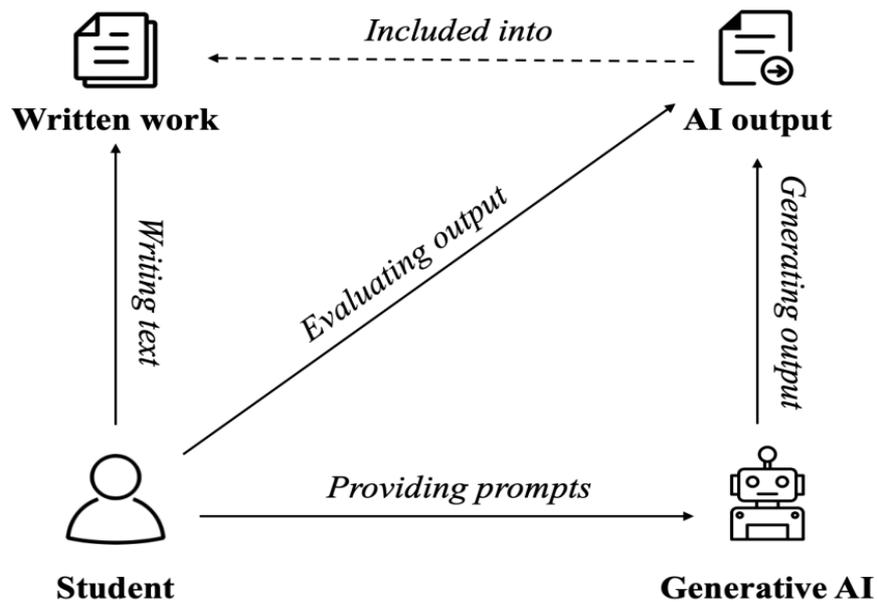

**Figure 1**. A 'machine-in-the-loop' approach to writing.

Research has been conducted to explore the role of a machine-in-the-loop in enhancing EFL writing. This research has primarily examined completed written works and associated outcomes. For example, Song and Song (2023) assessed the influence of ChatGPT on the IELTS writing skills of Chinese EFL students. Their findings showed notable enhancements in various writing skills such as organization, coherence, grammar, and vocabulary among students who were taught with the assistance of ChatGPT compared to those who underwent traditional instruction. In a similar vein, Tsai et al. (2024) found that EFL college English majors achieved significantly higher scores on essays revised with ChatGPT's assistance. All four dimensions of writing quality assessment saw significant improvements, with vocabulary showing the greatest improvement, followed by grammar, organization, and content.

Although studies indicate potential benefits from machine-in-the-loop writing, the benefits may depend on context and particular student interactions with AI. For instance, Woo et al. (2024b) examined how EFL secondary school students used AI-generated text and human-generated words in narrative compositions, revealing that the number of human words and AI-generated words significantly contributed to human-rated scores of the compositions. Besides, their study revealed that competent writers used AI-generated text strategically and in smaller quantities, but some less competent writers benefited from using more AI-generated text, suggesting that students' existing evaluation and editing skills influence effective use of AI. Escalante et al. (2023) compared the learning outcomes of university EFL students who received feedback on their 300-word weekly paragraph writing from ChatGPT versus human tutors. Their findings showed that the use of ChatGPT feedback did not lead to significant differences in learning outcomes compared to human feedback, indicating that ChatGPT feedback could be integrated in the EFL classroom as automated writing evaluation (AWE) software of EFL students' paragraphs without affecting learning outcomes.

We have a limited understanding of how discrete phases of writing with a 'machine-in-the-loop' may contribute to the overall quality of a written product. For instance, students may strategically formulate prompts, evaluate AI output, insert relevant AI output into the written work and modify those outputs and the students' own words in the written work. However, we do not know how any one of these phases significantly contributes to the value of the written product. Furthermore, we have limited understanding of student proficiency and timing of these phases. For example, a student may be less able to effectively evaluate or modify AI output for expository writing and take less time to do so than another student.

### *2.3. The Relationship Between Writing Process and Product*

Writing is underpinned by a cognitive process (Flower & Hayes, 1981). This involves the writer planning, drafting, and revising until a text meets its objective. The central focus of this process is the writer's ideas, their conversion into fitting words, and subsequently, their transcription into the evolving text. Van den Bergh et al. (2016) conceptualize the writing process as a functional dynamic system. In this system, cognitive activities, the task environment, and the writer's resources collaborate to enhance text quality. The authors emphasize the need to analyze the interplay of cognitive activities at specific points in time to better understand the writing process.

Studies have investigated the link between writing processes and writing performance (e.g., Bennett et al., 2020; Limpo & Alves, 2018; Sinharay et al., 2019). They have either examined the writing process at varying proficiency levels or attempted to forecast text quality based on process elements. Correlations have been found between students' writing process and the resultant product. For instance, the number of revisions is positively associated with improved writing quality (Wu & Schunn, 2021). Different types of revision activities have varying impacts; more substantive revisions, such as reorganizing content, are linked to higher text quality (Arias-Gundín et al., 2025). Additionally, students who engage in high-level revisions towards the end of the writing process tend to produce superior texts (Groenendijk et al., 2008).

In sum, although machine-in-the-loop writing can potentially enhance EFL students' argumentative and narrative compositions, effective interactions with generative AI for writing support appears nuanced and context-dependent. Furthermore, cognitive processes critically shape the final written product. However, previous studies have primarily focused on overall outcomes rather than understanding students' specific drafting and revising behaviors with AI-generated text. Given the existing challenges that EFL students face in expository writing and the limited research into AI use to support this genre, a gap exists in understanding EFL students' specific editing behaviors with AI-generated text in their expository compositions. Specifically, it remains unclear what types of edits students make, including insertions, deletions and modifications corresponding to expository writing units (e.g. title, heading and topic sentence) and the timing and extent of the edits to those units. It also remains unclear whether these edits demonstrably influence how people perceive the quality of final compositions.

Therefore, the study explores EFL students' specific edits and whether their edits enhance product quality. Importantly, we aim to shed light on students' edits from two perspectives, the *product* and the *process*. Investigating the *product* refers to our examining the students' final arrangement of AI-generated words and human words in the written work. Investigating the *process* refers to our exploring students' manipulation of AI-generated output while drafting and revising an expository composition, before submitting that composition for human scoring. Our exploration is guided by the following two research questions (RQs):

- *RQ1*: What are the characteristics of edits made by EFL secondary school students when integrating AI-generated text into their expository compositions, observed from both the process and product?
- *RQ2*: To what extent do these distinct editing characteristics, observed from both the process and product, predict the human-rated quality (in terms of content, language, organization and overall scores) of the final expository compositions?

## 3. Methodology

### *3.1. Research Context*

The study was conducted in seven Hong Kong secondary schools where English is taught as a foreign language and where students in EFL lessons are not taught expository essays but text types such as articles that include expository writing (Koh, 2015; The Curriculum Development Council, 2017). In Hong Kong mainstream education, individual schools admit students of similar academic capabilities (Lee & Chiu, 2017). Thus, to capture a diverse sample comprising a wide range of English proficiency levels and possible editing behaviors, we purposefully selected across the academic spectrum. Specifically, we selected three higher-achieving schools with students in the top 50% of academic percentiles and four lower-achieving schools with students in the bottom 50% of academic percentiles.

An English teacher at each school had enrolled their students in a free, two-hour workshop, which the first author designed and implemented at each school. With English as the medium of instruction, students reviewed the genre of expository writing and the process of writing an expository text type, specifically, an article. They were then introduced to prompt engineering strategies (Ouyang et al., 2022; White et al., 2023) to support expository writing. Students and researchers collaborated to practice crafting appropriate prompts for an example writing task. Specifically, they crafted a series of prompts to familiarize themselves with the text type, and then to plan, draft and revise the target text. For example, Figure 2 shows prompt writing tasks to plan the target text. The instructor and students assessed student prompts by students submitting prompts to Poll Everywhere (see Figure 3). The instructor and students discussed prompts and importantly tested them on the Platform for Open Exploration (POE) app, which is available online and aggregates state-of-the-art AI chatbots (see Figure 4). POE provides free access to American commercial chatbots such as ChatGPT and Claude that Hong Kong residents cannot directly access from OpenAI and Anthropic websites, respectively, because of geo-blocking.

**Task: Planning Content**
1. Work in pairs. Write a prompt that asks ChatGPT an information *question* about the topic. If you are satisfied with ChatGPT's output, write your prompt below.

_______________________________________________

2. Write a prompt that asks ChatGPT to *rewrite* its answer to your question in traditional Chinese language. If you are satisfied with ChatGPT's output, write your prompt below.

_______________________________________________

3. Now write a prompt for ChatGPT to *brainstorm* ideas for your topic. Be specific about how you want ChatGPT to *visualize* and organize its ideas. If you are satisfied with ChatGPT's output, write your prompt below.

_______________________________________________

**Task: Planning Language**
4. Write a prompt so that ChatGPT *brainstorms* useful vocabulary and grammar for writing this task. In addition, prompt ChatGPT to *reflect* on how it generated the vocabulary and grammar for the text, explaining its reasoning and assumptions. If you are satisfied with ChatGPT's output, write your prompt below.

_______________________________________________

**Task: Planning Organization**
5. Write a prompt so that ChatGPT will *generate* a writing plan or planning sheet for you to get an idea of how to organize your text. In addition, prompt ChatGPT so that it asks you four *questions* so that ChatGPT produces a better writing plan. Specify a format for *visualizing* its writing plan.

_______________________________________________

**Figure 2**. Tasks for writing prompts to plan a text.

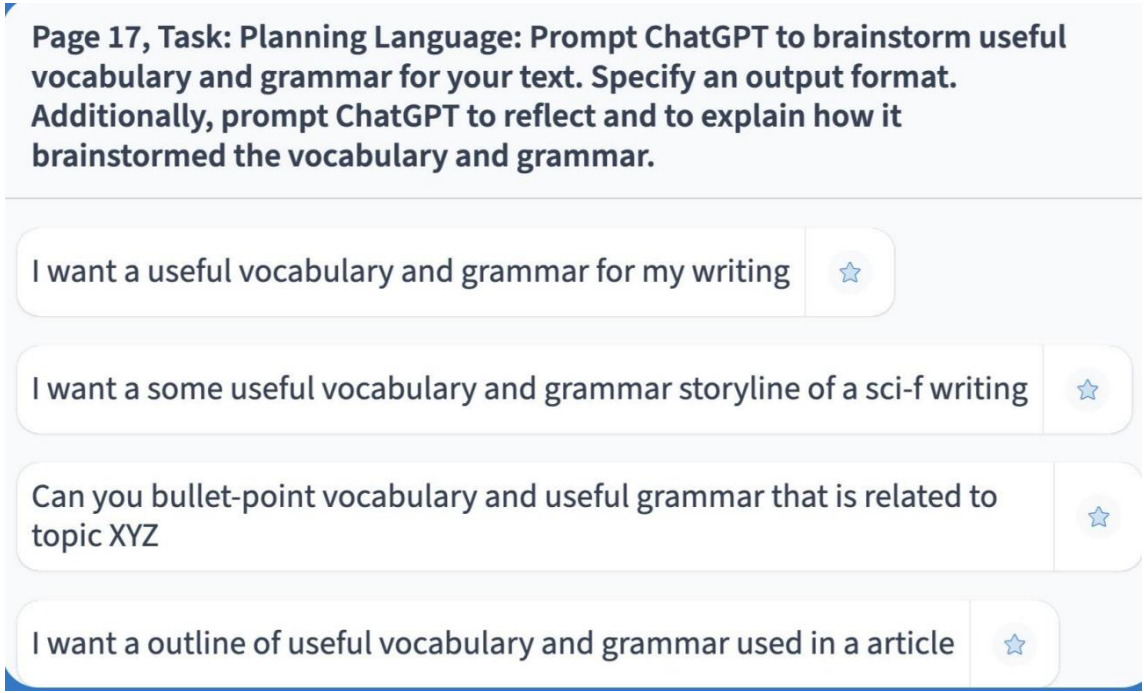

**Figure 3**. Student-submitted prompts on Poll Everywhere for a prompt that plans language.

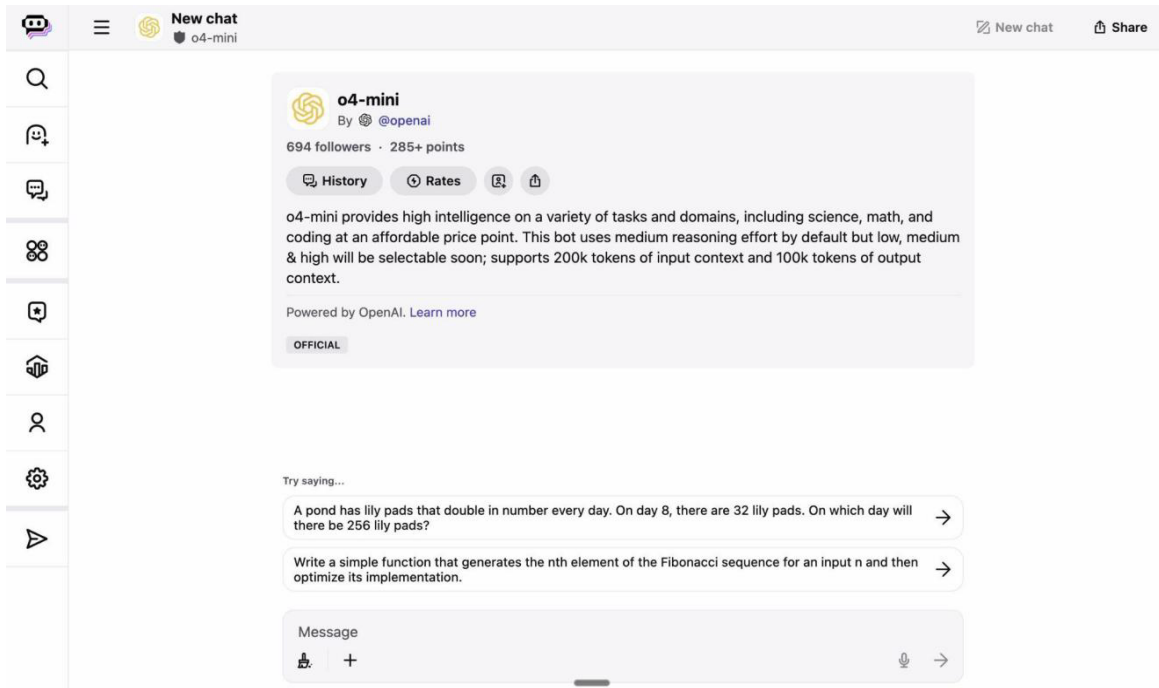

**Figure 4**. A chat window for the o4-mini chatbot on the POE app.

Students then independently attempted an expository writing task (see Figure 5) that is typically given at students' state-defined secondary school exit examination. To complete the task, students wrote an article of no more than 500 words, using their own words and AI-generated words from POE chatbots. They could use as much AI-generated text or as many of their own words as necessary. They differentiated between the two types of text by highlighting them in different colors (see Figure 6). Students wrote on Google Docs and were not required to finish the article at the workshop so that students could attempt the task at their own pace.

## Writing Tasks

A school must choose one of the following tasks for its students to answer.

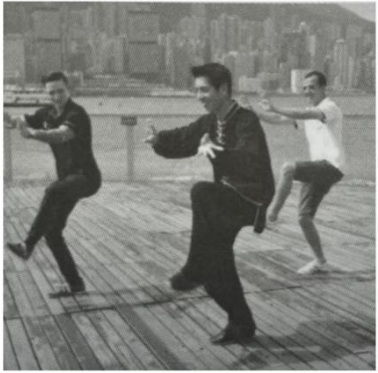

**Question 1**

**Learning English through Sports Communication**

While tai chi is a popular activity in Hong Kong, it is less known in some parts of the world.

Write an article for *International Travel* magazine introducing the benefits of tai chi to tourists.

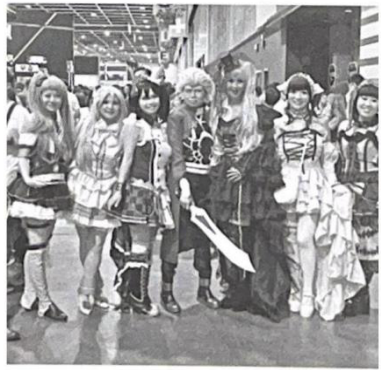

**Question 2**

**Learning English through Popular Culture**

Anime Expo, Hong Kong's biggest anime, manga and video game exhibition, was held at the Hong Kong Convention and Exhibition Centre last weekend. As a school reporter, you attended the event and interviewed some people dressed in cosplay.

Write an article for your school magazine.

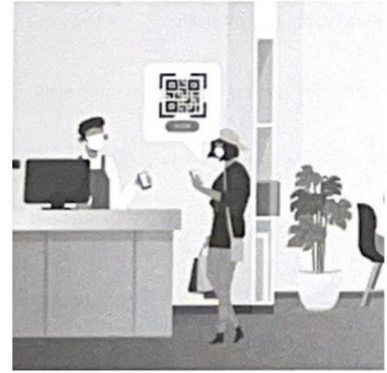

**Question 3**

**Learning English through Workplace Communication**

You work for *Restaurant Business* magazine. You interviewed a restaurant owner about his/her experiences of running a business during the pandemic.

Write a feature article for the magazine.

**Figure 5**. Three expository writing tasks.

### Resilience and Reinvention: The Untold Stories of Restaurants During the Pandemic
*Unveiling the Triumphs and Transformations of a Determined Industry*

**Introduction**

2020. The year that tested the world's resilience. Among the many sectors profoundly impacted by the global pandemic, the restaurant industry bore the brunt of unprecedented challenges. However, from the ruins rose a resilient exception, over which now a ray of dawn light shines after the merciless storm. In this exclusive interview with Mr Wu, the owner of Ho Ging Cha Chaan Teng, we had the privilege of taking in valuable information about his story, imbued with tenacity, valiance, and promise.

**The Tale of Mr Wu**

At the heart of our conversation, Mr Wu painted a vivid picture of the initial shockwave that reverberated through the industry. Forced closures, plummeting revenues, and the uncertainty of the future cast a dark cloud over their once-thriving establishment. Like many others, they were faced with a do-or-die moment necessitating adaptability and resilience.

**Hurdling over predicaments**

**Figure 6**. An article excerpt for Question 3 with AI-generated text in red.

*3.2. Participants*

Although seven Hong Kong secondary schools were purposefully sampled for this study, the students from those schools composed an opportunistic convenience sample as each school's teacher-in-charge was responsible for recruiting students for the workshop. Furthermore, we focused our analysis on compositions where students could engage in substantial drafting and revising with AI-generated text. Thus, we purposefully selected compositions from students who self-reported using at least 50% AI-generated text, excluding students who self-reported using minimal AI.

Thirty-nine students whose compositions met the selection criteria participated in this study. They were distributed across the seven schools, with 23 from the higher-achieving schools and 16 from the lower-achieving schools. In the pre-workshop questionnaire, 29 of them reported scoring more than 50 out of 100 marks in their last English writing exam, and 11 reported having written an article in English before. As for generative AI experience, 26 reported having used ChatGPT, and 10 reported having used ChatGPT specifically to complete English language homework. The variation in language proficiency and previous generative AI experience ensured the sample's diversity and allowed for a relatively comprehensive understanding of students' editing. At the workshop, all the students were informed of the study and their rights and voluntarily agreed to participate.

*3.3. Data Collection*

To answer the research questions from *process* and *product* perspectives, we planned a convergent design (Creswell & Clark, 2018), where we would analyze two sets of equally important data and compare results. Table 1 elaborates our research design in terms of data sources, details, purposes and analytical methods. For the *process* perspective, we aimed to collect data of students inserting AI-generated output into a document, and deleting or modifying any AI-generated text in the document. We also aimed to explore the temporal nature of these edits. To capture the editing and timing, we requested students to record their screens while writing the task during the workshop. Importantly, since students were not time-bound to complete the writing task at the workshop, our screen recordings compose a convenience sample of students' editing and a recording may not have captured any or all of a student's editing of a composition. In that way, screen recordings were collected from 39 participants; however, only 25 recordings were eligible for analysis, as 10 participants did not submit their recordings and 4 participants' recordings showed no AI-generated text edits. For the *product* perspective, we collected data of students' final arrangement of AI-generated text and their own words in the document. Students wrote on Google Docs and shared with the researchers. In the event that students wanted additional time to write their compositions, we reviewed students' compositions two weeks after each workshop.

**Table 1.** Data sources, details, purposes and analytical methods.

| Process writing stage | Data source | Data details | Purpose | Analytical methods |
|---|---|---|---|---|
| Drafting | *Process-oriented:*<br><br>Screen recordings (n=25) | - Total length:<br>  10 hours, 34 seconds<br>- Average length:<br>  24 minutes<br>- Range of length:<br>  1 minute, 6 seconds to<br>  35 minutes, 22 seconds | To identify students inserting, deleting and replacing AI-generated text on written compositions before submission | - Qualitative coding<br><br>-Descriptive statistics<br><br>-Temporal sequence analysis<br><br>-Human-rated scoring<br><br>- Multiple linear regression analysis |
| Revision | | | | |
| Submission | *Product-oriented:*<br><br>Written compositions (n=39) | - Word count range:<br>  234 to 500 words<br>- Average: 441 words<br>- Median: 476 words | To explore students' final arrangement of AI-generated text and their own words in written compositions | |

## *3.4. Data Analysis*

### *3.4.1. Qualitative Coding and Descriptive Statistics*

We answer RQ1 qualitatively, performing a content analysis (White & Marsh, 2006) by implementing a hierarchical coding scheme, and presenting descriptive statistics. To prepare data for coding, we noted to which school, form level, and student each written composition and screen recording belonged. For the latter, we also noted the total runtime.

To explore the characteristics of edits made by EFL secondary school students, we focused on the target and scope of their editing actions. First, we designed descriptive, provisional codes for the target of students' editing actions in terms of key structural components of expository writing. In this way, the first and fifth authors, EFL writing teachers in Hong Kong secondary schools, reviewed Hong Kong secondary school EFL textbooks (Nancarrow & Armstrong, 2024; Potter et al., 2023). They operationalized six expository writing features or units that were found in both textbooks' instructions on how to write an article: (1) title, (2) heading, (3) body paragraph topic sentence, (4) body paragraph supporting sentence, (5) introductory paragraph and (6) concluding paragraph. Second, we designed another set of descriptive, provisional codes to detail the scope of each edit in relation to expository writing units. Drawing on approaches to analyzing the length of text production units (Lu, 2010) and refining the categorization system used by Woo et al. (2024b), we designed three codes to differentiate between minor, localized edits and more substantial, unit-level edits: (1) *short edits*: less than an expository writing unit (e.g., a word in a title; words in a topic sentence); (2) *medium edits*: exactly one unit (e.g., a topic sentence; a supporting sentence; a title), representing manipulation of what is considered a structurally complete, predefined element of expository text; or (3) *long edits*: exceeding one unit (e.g., a topic sentence plus additional words in a supporting sentence; words that span across several expository units). Categorizing edit scope relative to an expository unit allows for a systematic description of the magnitude or extent of students' edits. By observing patterns of edit scope across different expository units, we can gain insight into students' engagement and knowledge of expository writing.

We applied the expository unit codes and the edit scope codes simultaneously for each student edit that we identified during the initial coding phase. See Appendix 1 for the final coding scheme. We applied the coding scheme to the compositions first. Since students may compose final compositions with chunks of AI-generated text that are indistinguishable from each other, we operationalized student edits of AI-generated text as student insertions of their own words between chunks or instances of AI-generated text. Thus, we coded every *chunk* or instance of a student's own words for its expository writing unit and its edit scope. Furthermore, for each composition, we analyzed the basic structure and organization in terms of the number of human words and AI words, and the total number of words. We presented frequency counts of human chunks as descriptive statistics.

Second, we applied the coding scheme to the screen recordings. We observed each recording, coding chronologically each instance where a student directly manipulated AI-generated text in the composition. We coded each AI-generated text edit for its expository writing unit and its edit scope, along with the timestamp indicating when the edit began. We coded according to how students self-reported syntactic units such as sentences and paragraphs with full stops and line breaks, respectively. We would validate high inference coding, such as when students initially wrote on Google Docs, with what students subsequently wrote in their docs.

The first author coded all composition and screen recording data and wrote observational notes all on Google Sheets. To enhance the trustworthiness of the coding, the first author compiled a codebook and implemented intercoder agreement procedures. For composition coding, the fifth author randomly selected and coded 15% of the final compositions (n=6). The first and fifth authors analyzed their independent coding with the expository unit codes and edit scope codes, and agreed on 94% of coding instances. After discussion, the two coders resolved all discrepancies and achieved 100% agreement. For screen recording coding, the second author randomly selected and coded 15% of the screen recordings (n=4). The first and second authors compared the coding, resolved a discrepancy by revising a code example in the codebook and achieved 100% agreement.

*3.4.2. Temporal Sequence Analysis*

To explore the temporal characteristics of students' editing during the AI-assisted writing process, we conducted temporal sequence analysis with data coded from the screen recordings. Since the recordings captured the drafting phase during the workshop, their lengths and coverage varied among the students. To increase comparability across different students, we first aligned each student's editing sequence by setting the timestamp of their first edit as the starting point (00:00). We then applied optimal matching (OM) to cluster students based on the order and transition of their edits by expository unit. OM computes the dissimilarity between two sequences by determining the minimum cost required to transform one sequence into the other through a series of insertions, deletions, and substitutions (Gabadinho et al., 2011). To specify the temporal characteristics of each cluster, we visualized students' edits using time-aligned scatterplots, which plotted each edit by its timestamp and expository unit. The combined approach helped reveal both editing patterns in terms of sequence and editing rhythm in terms of time. All sequence analysis, clustering, and visualization were conducted using R with the TraMineR and ggplot2 packages.

*3.4.3. Human-rated Scoring*

Prior to scoring, the compositions were anonymized so that scorers would not know who wrote a composition and what words in a composition were AI-generated text. Each composition was scored for content, language and organization, according to the marking scheme (see Appendix 2) that EFL teachers commonly use to score compositions for Hong Kong's secondary school exit examination. The highest possible score for each criterion was 7, so the highest possible total score was 21. Two experts independently scored each composition. One of the experts was the EFL teacher-in-charge from a participating school, and the other was the first author who was also an EFL teacher-in-charge. These experts' scores were averaged to arrive at a final content, language, organization and total score for the composition.

*3.4.4. Multiple Linear Regression Analysis*

We answer RQ2 by performing multiple linear regression (MLR) analyses using the quantified editing characteristics identified in RQ1 and human-rated composition scores. A regression model was attempted respectively for four score items, namely content, language, organization and total scores, using the same set of independent variables that reflected students' editing. For the *process* perspective (n=25), we included as potential predictors the total runtime and the numbers of AI-generated text edits coded from the screen recordings. For the product perspective (n=39), we selected the numbers of AI words, human words and human chunks coded from the written compositions. We separated the MLR analyses for the two perspectives, because encompassing too many independent variables at one time can lead to issues such as multicollinearity, overfitting and model complexity (Babyak, 2004). Statistical Package for the Social Sciences (SPSS, version 26.0) was employed to perform the analyses using the enter and backward methods. Partial correlation values were calculated to suggest how a single variable may predict a score item in a constructed model after controlling the effect of other variables.

## 4. Results

Per Table 1, we had collected 39 compositions. The percentage of AI-generated text found in these compositions ranged from 52.8% to 99.7% with an average of 83.7% and a median of 89.5%. The number of instances that students used their own words to edit these largely AI-generated compositions ranged from 1 to 34 with an average of 6.7 and a median of 5. In the 25 screen recordings, we observed most students interacting with a chatbot on POE as opposed to interacting with several chatbots on the app. Only one student consulted a chatbot outside the POE app.

## 4.1. Characteristics of Students' Edits (RQ1)

### 4.1.1. Expository Units and Edit Scope

Table 2 presents the frequency count of student edits organized by expository units and edit scope. It also compares these student edits between process and product perspectives. We find a similar total number of edits for the process (n=266) and the product (n=261) perspectives. The most frequently edited expository unit from the process perspective is the introductory paragraph (n=60) and the least frequently edited is the concluding paragraph (n=15). From the product perspective, the most frequently edited expository unit is the supporting sentence (n=91) and the least is the heading (n=6). Besides, we had observed instances of students editing unconventional or inappropriate units (e.g., a letter greeting; a letter signature; a writing task prompt; an AI-generated outline; a chatbot prompt). We coded these unconventional units under the category, "Other." In terms of edit scope, student edits are most frequently short chunks from both the process (n=155) and product (n=202) perspectives.

**Table 2.** Number of student edits specified by perspectives and expository units.

| | Short Chunks | | Medium Chunks | | Long Chunks | | Total by expository features | |
|---|---|---|---|---|---|---|---|---|
| Perspective | Process | Product | Process | Product | Process | Product | Process | Product |
| Title | 3 | 4 | 10 | 5 | 11 | 3 | 24 | 12 |
| Heading | 6 | 0 | 10 | 2 | 7 | 2 | 23 | 4 |
| Topic S | 19 | 64 | 0 | 7 | 21 | 10 | 40 | 81 |
| Supporting S | 44 | 65 | 8 | 12 | 6 | 14 | 58 | 91 |
| Intro P | 51 | 41 | 7 | 1 | 2 | 0 | 60 | 42 |
| Concluding P | 12 | 27 | 3 | 1 | 0 | 0 | 15 | 28 |
| Other | 20 | 1 | 13 | 2 | 13 | 0 | 46 | 3 |
| Total by edit scope | 155 | 202 | 51 | 30 | 60 | 29 | 266 | 261 |

*Note*: the number of product edits refers to instances of human chunks from 39 written compositions; and the number of process edits refers to instances of AI-text manipulation from 25 screen recordings.

Figures 7 and 8 show the frequency count of student edits by expository units and student distribution for the process and product perspectives, respectively. Comparing the number of students who edited an expository unit from the process perspective, the largest number of students (n=16) edited the introductory and the smallest (n=5) the heading. From the product perspective, the largest number of students (n=27) edited the topic sentence and the smallest number (n=2) edited the heading.

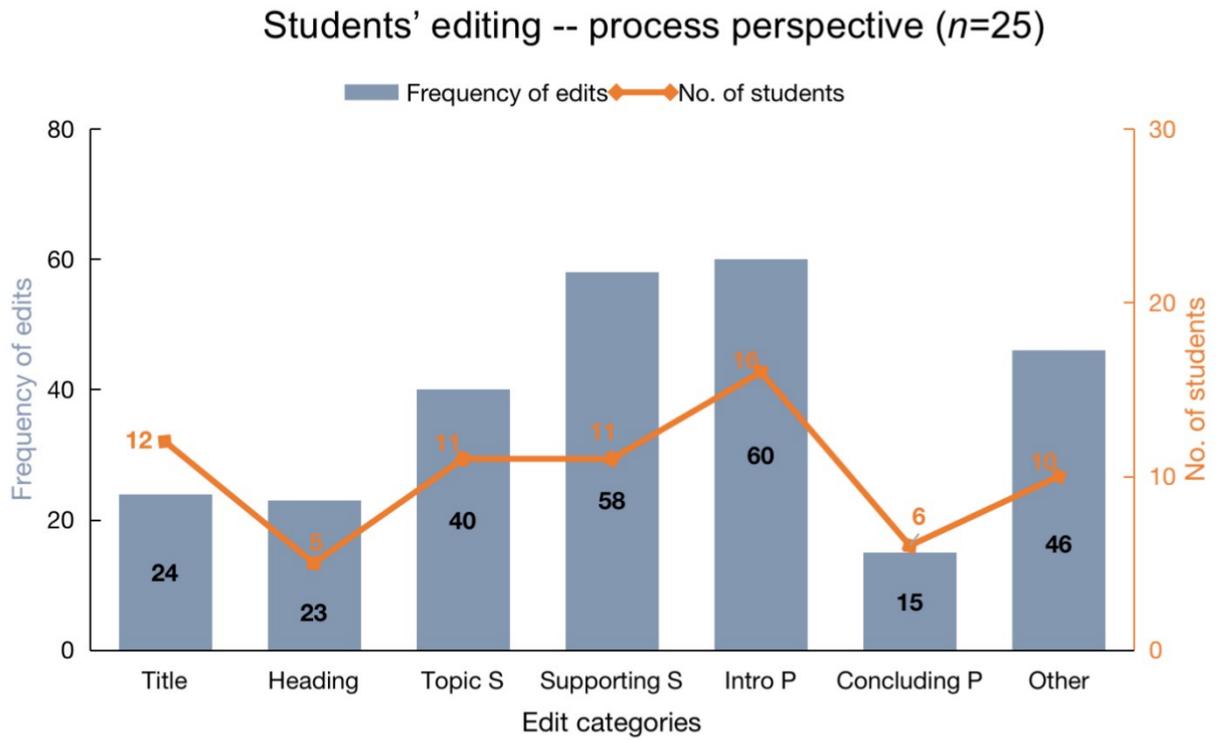

**Figure 7**. Frequency of edits and student distribution from the process perspective.

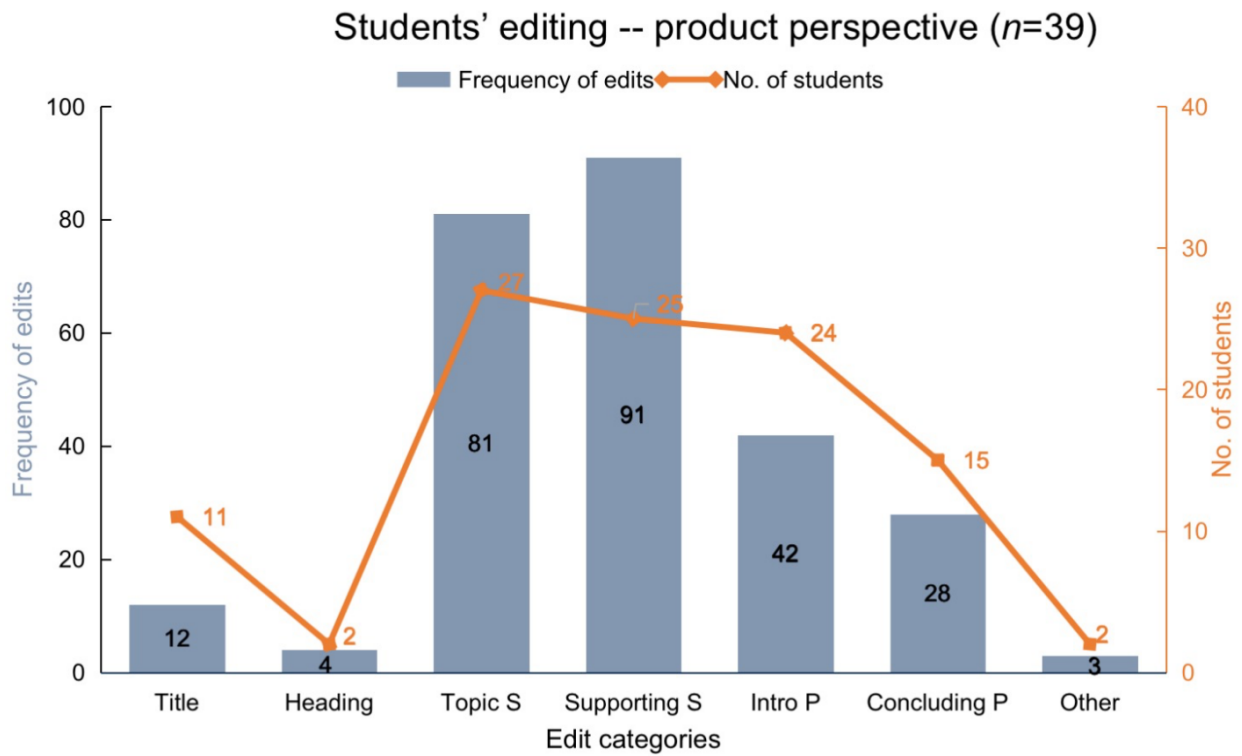

**Figure 8**. Frequency of edits and student distribution from the product perspective.

*4.1.2. Editing patterns from temporal sequence analysis*

Two distinct clusters of editing trajectories were identified under the OM approach. Figure 9 presents the most frequent sequences within each cluster, indicating students' editing patterns by expository unit and order. Both clusters share a feature of students starting their editing with expository units found at the beginning of an article on the top of a page. Cluster 1 (n = 19) showed relatively high sequence diversity. The cluster's students followed a variety of paths when editing AI-generated text, but their sequences often included repeated edits in other (O), introductory paragraph (IP), and topic sentence (TS). For example, one student made 11 consecutive short edits to a long heading above the title with unknown purpose (coded O). The shared feature points to an introduction-oriented editing pattern, where the students repeatedly refined or re-evaluated early sections of a text before moving on. On the other hand, Cluster 2 (n = 6) showed more consistency among its students. The sequences in this cluster typically started with title (T), other (O), and introductory paragraph (IP) edits, and prominently featured recurrent edits to topic sentence (TS) and supporting sentence (SS). This indicates a body-oriented editing pattern, where the students focused much on paragraph-level content after initial attention to the beginning units.

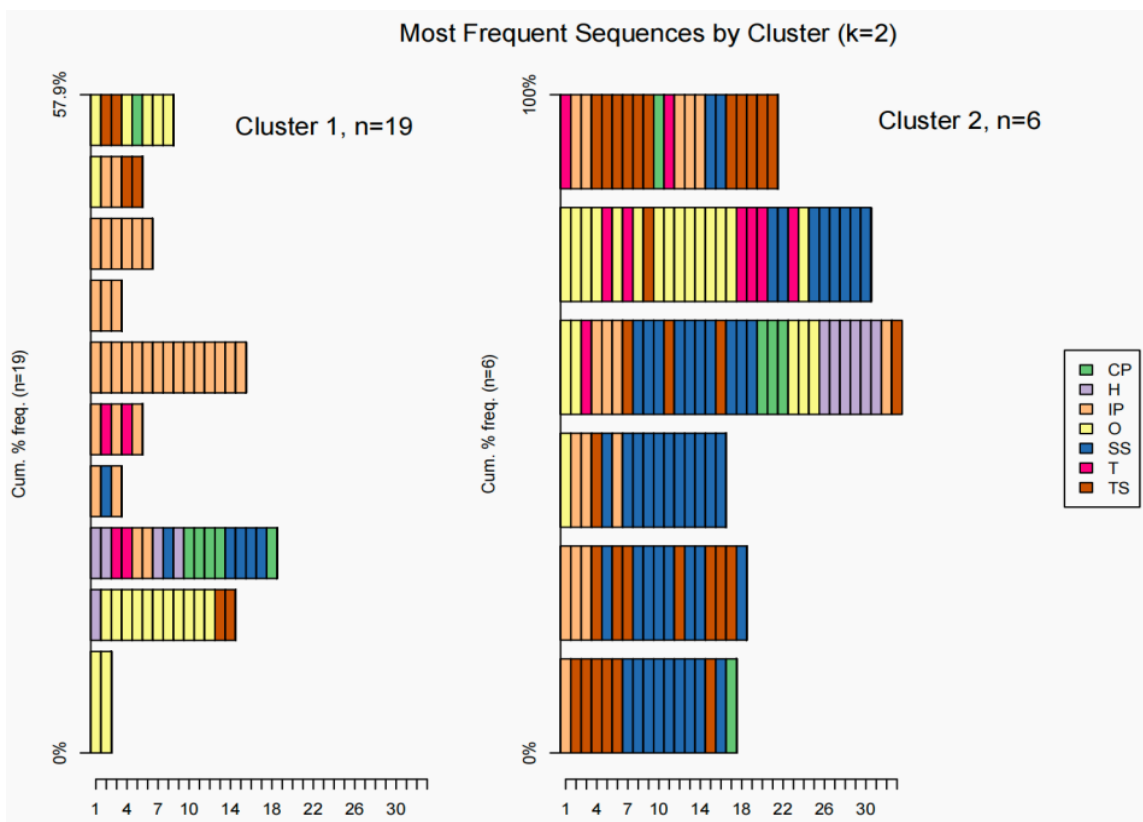

**Figure 9.** Most frequent edit sequence in two identified clusters.

Figure 10 visualizes students' editing trajectories by cluster with time-align scatterplots. Each dot represents an individual edit, its position reflecting the adjusted timestamp (x-axis) and the expository unit being edited (y-axis). The scatterplots generally corroborate the sequence-based findings, but additionally reveal when and how persistently students engaged with different text parts over time. First, examining the initial onset of editing (0 second), students from both clusters placed entry points in title (T), introductory paragraph (IP), and other (O). Then differences emerged in the first 500 seconds, during which Cluster 1 students focused intensively on editing the beginning units, whereas Cluster 2 students started to move on to body paragraphs through topic sentence (TS) edits. From 500 to 1000 seconds and afterwards, the attention to the beginning units continued among Cluster 1 students, with only a few progressing to edit topic sentence (TS), supporting sentence (SS) and concluding paragraph (CP). In the same time span, Cluster 2 students centered on the body paragraphs with plentiful TS and SS edits.

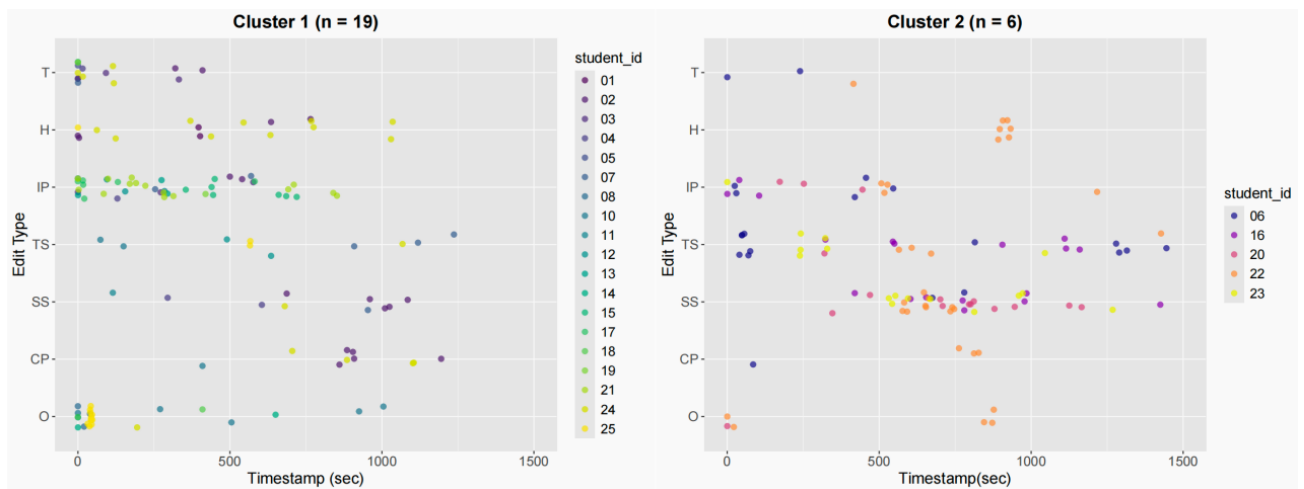

**Figure 10.** Time-aligned scatterplots of students' editing behaviors by cluster.

Though confined to the initial writing process, the editing trajectories based on temporal sequence analysis suggested in Figures 9 and 10 jointly suggested two possible editing patterns: one distributed yet top-heavy editing rhythm involving concentrated edits to the opening (Cluster 1), and another concentrated and temporally compact editing pattern characterized by relatively early engagement with body part (Cluster 2).

*4.2. Impacts of Student Editing on Composition Quality (RQ2)*

Table 3 summarizes the human-rated scoring for content, language, organization and total scores for compositions, further organized by the process compositions (n=25) and the product compositions (n=39). The scores are the average of two experts' scores. Seven is the highest possible score for content, language and organization criteria. Twenty-one is the highest possible total score which one student achieved in the process cohort and two in the product.

**Table 3.** Human-rated scoring for student compositions

|  | **Content** | | **Language** | | **Organization** | | **Total** | |
|---|---|---|---|---|---|---|---|---|
|  | **Process** | **Product** | **Process** | **Product** | **Process** | **Product** | **Process** | **Product** |
| Average | 5.1 | 4.8 | 5.5 | 5.3 | 4.8 | 4.6 | 15.4 | 14.6 |
| Standard Deviation | 1.2 | 1.3 | 1.0 | 1.1 | 1.0 | 1.1 | 3.1 | 3.3 |
| Minimum | 2.0 | 0.5 | 3.0 | 2.0 | 2.5 | 1.5 | 7.5 | 4.0 |
| Maximum | 7.0 | 7.0 | 7.0 | 7.0 | 7.0 | 7.0 | 21.0 | 21.0 |

*4.2.1 MLR Findings From the Screen Recordings*

We first explored the relationship between students' editing and composition quality from the *process* perspective by developing regression models for the four score items, the independent variables being the total runtime and the seven categories of students' edits by expository units. Since no statistically significant models were obtained with the enter method, we applied the backward elimination method to identify relevant predictors based on adjusted *R*-squares and *p*-values. A model was constructed for language score with statistical significance, $F(3,21) = 3.153$, $p = .046$, and a model for total score was constructed with marginal statistical significance, $F(2, 22) = 2.823$, $p = .081$. However, no such models could be established for content score $F(2, 22) = 2.440$, $p = .110$, and organization score $F(1,23) = 2.407$, $p = .134$. The constructed models had no significant issues of multicollinearity or autocorrelation, as we calculated Variance Inflation Factor (VIF), for which all variables had a value less than 10, and Durbin-Watson statistics, which were found to be around 2.

Table 4 shows a summary of partial correlation results with respective model fit indices in the eight-variable analyses. The numbers of Title, Heading, and ConcludingP edits were included in the language score model, accounting for 31.1% of the variance. The numbers of Heading and ConcludingP edits were again found in the total score model, accounting for 20.4% of the variance. Partial correlation values revealed that Heading edits positively predicted the language and total scores, while ConcludingP edits negatively influenced them.

Additionally, we conducted five-variable MLR analyses with total runtime and frequently used expository unit and edit scope combinations. Four combinations found in at least one-third of the recordings were included: Title(Medium), Topic S(Long), Supporting S(Short), Intro P(Short). However, the analyses yielded no statistically significant model for any of the score items, showing minimal explanatory power of these combinations. Full regression results for each model in both eight-variable and five-variable analyses, including standardized coefficients (β), t-values, p-values and VIFs, can be found in Appendix 3.

**Table 4**. Summary of partial correlation in the eight-variable MLR analyses (process-oriented, $n$=25)

|  | **Content score** | **Language score** | **Organization score** | **Total score** |
| --- | --- | --- | --- | --- |
| Total runtime | - | - | - | - |
| No. of Title edits | - | 0.259 | - | - |
| No. of Heading edits | 0.417* | 0.470* | - | 0.447* |
| No. of Topic S edits | - | - | - | - |
| No. of Supporting S edits | - | - | -0.308 | - |
| No. of IntroP edits | - | - | - | - |
| No. of ConcludingP edits | -0.403† | -0.491* | - | -0.419* |
| No. of Other edits | - | - | - | - |
| Model $p$-value | 0.11 | **0.046** | 0.134 | 0.081 |
| $R$-squared value | 0.182 | 0.311 | 0.095 | 0.204 |
| Adjusted $R$-squared value | 0.107 | 0.212 | 0.055 | 0.132 |

*Note*: "*" means the partial correlation value is significant with $p < .05$, "**" means $p < .01$, and "†" means $.05 < p < .10$, marginally significant.

### 4.2.2. MLR Findings From the Written Compositions

To understand the contribution of students' editing to composition quality from the *product* perspective, we constructed MLR models for the four score items with potential predictors extracted from the written compositions. First, we performed nine-variable MLR analyses using the enter method, where the numbers of AI words, human words, and the seven expository unit categories were simultaneously included as predictors. A statistically significant model was constructed respectively for content score, $F(9,29) = 3.756$, $p = .003$, language score, $F(9,29) = 3.440$, $p = .005$, organization score, $F(9,29) = 2.783$, $p = .018$, and total score, $F(9,29) = 3.662$, $p = .004$. The nine variables accounted for 46.3% to 53.8% of variance in the score items. The models had no significant issues of multicollinearity or autocorrelation.

Table 5 shows a summary of partial correlation results with respective model fit indices in the nine-variable analyses. All score items were negatively correlated with the number of Heading chunks and the number of Other chunks. Figure 11 illustrates an Other chunk example as the writer began the article with a letter greeting comprising AI words in red and the writer's own words in black. The number of AI words was moderately correlated with every score item in a positive way, while the number of Human words influenced content, organization and total scores with marginal significance. The remaining variables, including the numbers of Title, Topic S, Supporting S, Intro P and Concluding P, were not found to independently predict the score items with statistical significance. In other words, each of them did not have a strong unique effect on scores after the other variables were controlled.

<span style="color:red">Dear</span> esteemed readers,

<span style="color:red">I hope this letter finds you in good health and high spirits. As an avid traveler and a passionate advocate for cultural exploration, I</span> would like to share an extraordinary experience that awaits <span style="color:red">you in the heart of Hong Kong - the ancient art of Tai Chi.</span>

**Figure 11**. An Example of an Other Chunk Comprising AI Words and Human Words

**Table 5**. Summary of partial correlation in the nine-variable MLR analyses (product-oriented, *n*=39)

|  | **Content score** | **Language score** | **Organization score** | **Total score** |
|---|---|---|---|---|
| No. of AI words | 0.417* | 0.457* | 0.375* | 0.436* |
| No. of Human words | 0.331† | 0.322 | 0.341† | 0.348† |
| No. of Title | -0.148 | -0.172 | -0.119 | -0.154 |
| No. of Heading | -0.521** | -0.519** | -0.497** | -0.533** |
| No. of Topic S | 0.290 | 0.204 | 0.159 | 0.234 |
| No. of Supporting S | -0.062 | -0.043 | -0.056 | -0.057 |
| No. of Intro P | -0.006 | -0.101 | -0.116 | -0.075 |
| No. of Concluding P | 0.098 | -0.074 | 0.077 | 0.088 |
| No. of Other | -0.348† | -0.350† | -0.395* | -0.381* |
| Model *p*-value | **0.003** | **0.005** | **0.018** | **0.004** |
| *R*-squared value | 0.538 | 0.516 | 0.463 | 0.532 |
| Adjusted *R*-squared value | 0.395 | 0.366 | 0.297 | 0.387 |

Note: "*" means the partial correlation value is significant with $p < .05$, "**" means $p < .01$, and "†" means $.05 < p < .10$, marginally significant.

We also performed six-variable MLR analyses with expository unit and edit scope combinations that appeared in at least one-third of the compositions. The six independent variables were the numbers of AI words, Human words, Topic S(Short), Supporting S(Short), Intro P(Short), and Concluding P(Short). With the backward elimination method, the final models for content, language, and total scores included AI words, Human words, and Topic S(Short), while the model for organization score included merely AI words and Human words with marginal statistical significance. Partial correlation results of AI words and Human words were consistent with the nine-variable analyses, but Topic S (Short) chunks, albeit in three models, did not turn out to independently affect these scores. Full regression results for each model in both nine-variable and six-variable analyses, including standardized coefficients (β), t-values, p-values and VIFs, can be found in Appendix 4.

## 5. Discussion

### *5.1. Characteristics of Students' Edits*

We observed 266 edits in 25 screen recordings and 261 edits in written compositions, with the majority of edits being short chunks. The large number of edits in both process and product indicate students had put considerable effort into editing, and suggests students preferred minimal, localized modifications rather than expository-unit-level modifications. This effort may get overlooked in EFL writing assessments that focus exclusively on the product, or the preponderance of AI-generated text in the product.

Students most frequently edited the introductory paragraph during the writing process but most frequently edited supporting sentences in the final product. These findings can indicate students changing drafting behaviors from iterative cycles of drafting and revising. They can also indicate that students are writing sequentially from the first paragraph onwards; in that way, students most frequently edit the supporting sentence because this expository unit is typically the most numerous in an expository text.

Heading units received little attention compared to other expository units and were edited by the fewest number of students in both the process and the product. Additionally, a few students drafted and revised unconventional units for expository writing. These findings may reflect various factors including unfamiliarity with an article text type as only 11 students had self-reported previously writing an article in English. The findings may also reflect students' limited knowledge of a full range of expository writing features. That could be because in the Hong Kong secondary school EFL curriculum, students are typically not taught to read and write expository essays but are taught narrowly defined text types that can include expository writing (Koh, 2015; The Curriculum Development Council, 2017).

From the temporal sequence analysis, we found students started their editing with expository units found at the start of an article on the top of a page. However, two distinct editing patterns emerged: Cluster 1 points to an introduction-oriented editing pattern, where the students spent much time repeatedly refining or re-evaluating beginning units of a text before moving on. Cluster 2 points to a body-oriented editing pattern, where the students move on more quickly from the beginning units and make extensive edits in the body part. Besides, editing rhythms varied considerably within the recorded writing time. These findings reflect different preferences for making use of drafting and revision time for expository writing. They also reflect different strategies emerging from what appears to be sequential writing from the top of an article to the bottom. That students are writing sequentially, albeit with different time-based strategies, may point to the heavy cognitive load of writing with a machine-in-the-loop (Woo et al., 2024c). Students may not be attempting other strategies, for instance, developing body paragraphs and a conclusion first and an introduction last because of heavy cognitive demands from drafting and revising with AI-generated text. Similarly, if students are merely writing sequentially, they may not be paying sufficient attention to the composition's overall organization (Sasaki & Hirose, 1996).

*5.2. Impacts of Student Editing on Composition Quality*

The MLR analyses of screen recordings included potential variables related to the direct manipulation of AI-generated text while composing an article. The MLR analyses identified fine-grain patterns of student editing that affected the quality of their compositions. First, students who edited heading units during the writing process achieved higher language and total scores. However, students who edited concluding paragraphs during the process received lower language and total scores. The former behavior suggests that students who edit a heading may possess more knowledge not only of language but also of expository writing units, because they are putting effort and time into drafting and revising a cohesive feature for body paragraphs. In contrast, the latter behavior suggests students in the span of a screen recording have finished drafting and revising an article and may not be revising language effectively.

Other process perspective MLR results showed limited predictive power for most editing variables. That suggests students' observable AI-generated text editing behaviors do not uniformly translate to human-rated composition scores. This resonates with the descriptive statistics findings, indicating many students spent time and effort editing their compositions but not all students composed high-quality final compositions. Our findings contrast those from Tsai et al. (2024) who found their EFL students achieved significantly higher scores on compositions revised with ChatGPT. The differences in these two studies' findings could be attributed to different student samples and text types, and importantly, different procedures for using AI in the revision process. However, like Tsai et al.'s (2024) study, we note that AI-assisted revision does not directly reveal a student's writing competence.

The MLR analyses of final compositions included potential variables related to the final arrangement of human words embedded in AI-generated text. First, all score items were negatively correlated with the number of heading chunks; and the number of other chunks was a statistically significant negative predictor for organization and total scores. However, these negative correlations should be interpreted with caution. This is because heading and other chunks were respectively found in 2 out of 39 students, leading to extreme data sparsity that limits the robustness of the identified negative correlations. As such, we may regard them as an initial signal rather than conclusive evidence. Besides, the organization score models had the lowest adjusted R-square among the four, suggesting the influence of students' editing on their organization performance may be qualified compared to content or language.

The nine-variable MLR analyses also found the number of AI words positively predicted all score items with statistical significance. The number of human words was positively correlated with two scoring items and total score with marginal significance. These results are similar to Woo et al.'s (2024b) finding that the number of human words and AI-generated words significantly contributed to the scores of students' narrative compositions. The mechanisms for an increasing number of AI words benefiting student scores, proposed in their study, may be applicable to interpreting our findings. Specifically, an increasing number of AI words may benefit less competent writers in the drafting stage and more competent writers in the revision stage. Furthermore, while short chunks constituted the majority of student edits of various expository units, they exerted limited influence on the quality of final compositions as suggested by the supplementary six-variable MLR analyses. This suggests the scope of edits alone does not determine their effectiveness in terms of human-rated scores.

In sum, the convergence of process and product results show that students invest much effort in editing AI-generated text yet that effort does not necessarily translate into quality improvements, suggesting a gap between their engagement and their writing competence. Moreover, that AI word count positively predicts scores while most editing behaviors show limited impact suggests students may be more effective at strategically incorporating AI-generated text than developing their own ideas.

*5.3. Implications*

The observed disconnect between considerable student editing effort and limited improvements in composition quality highlights that AI complements EFL students' writing competence. AI is not a panacea for underlying writing deficiencies. Therefore, following Hyland's (2007) genre-based framework, educators should systematically teach expository principles such as structuring introductions, crafting clear topic sentences and developing coherent paragraphs with appropriate headings. Furthermore, following process-based writing principles (Flower & Hayes, 1981), students should be taught to systematically plan, draft and revise. For example, during a planning stage students could use graphic organizers to visualize logical flow and consider alternatives to the common sequential drafting pattern our temporal analysis revealed.

After foundational writing knowledge, machine-in-the-loop writing should be introduced with an aim for students to be critical and active evaluators, not passive consumers of AI-generated text. In that way, teachers can model effective machine-in-the-loop writing, for example, to generate and compare alternative topic sentences, to check paragraph coherence against genre criteria, or to explore multiple conclusions. Moreover, students can compare human-written and AI-generated expository texts, identifying strengths and weaknesses in coherence, stylistic appropriateness, etc. to develop analytical skills for expository writing with a machine-in-the-loop. Students could also learn to integrate AI-generated text and their own words at different edit scopes. In addition to scaffolded practice, collaborative learning activities can support students who are reluctant to use AI or face technical barriers. Our recommendations for leveraging generative AI to enhance expository writing skills contribute to practices for teaching children to write expository texts (Fang, 2014).

Ultimately, EFL educators should frame AI as a tool that complements existing writing competence, and that independent analytical and writing capabilities are essential with or without AI assistance. That framing should correspond with developments in assessment practices that value the writing process, including students' interaction with and critical evaluation of AI-generated text alongside the final product. For example, educators could implement portfolio assessments that document students' planning, AI-generated text interactions and revisions, and reflections. That formative assessment aligns with Van den Bergh et al.'s (2016) emphasis on understanding dynamic cognitive activities and better captures the learning occurring during machine-involuted writing.

### *5.4. Limitations and Future Research*

Although the school sample is robust for the Hong Kong context, the study is small-scale in terms of the participant numbers. Given the relatively small sample size, the MLR analyses were largely exploratory in nature and likely underpowered. Thus, the results should be interpreted with caution, and further research with larger samples is needed to confirm these findings. Future studies should expand the sample to a greater number of students and to all compositions comprising AI-generated text, including those written with less than 50% AI-generated text, to find associations between editing of that text and human-rated scores with improved statistical power.

Another study limitation is the length of screen recordings which only captured the beginning of students' writing-with-a-machine in the loop in a social setting. Students' may have performed more editing before collection of their final compositions two weeks after a workshop. Moreover, students' patterns may change if given a time-bound writing task where students must complete a text in a limited time. Therefore, future research may attempt to record and observe an EFL student's entire process from planning to submitting the final composition.

More data could be collected about students' expository writing knowledge so as to validly interpret student editing behavior in terms of their existing knowledge. Adding data sources including students' and teachers' perceptions of AI use in expository writing could also strengthen analysis.

Finally, we require further research to identify salient patterns of editing AI-generated text that lead to improved outcomes. That could start with specifying behavioral features of AI-generated text edits during the drafting and revising process and then using MLR analyses to explore relationships between a student's editing behaviors and his/her final composition quality.

## 6. Conclusion

This study has explored EFL students' expository writing with a machine-in-the-loop from process and product perspectives. It has evidenced different characteristics of students' editing of AI-generated text in their expository writing and how these characteristics may or may not enhance product quality in terms of human-rated scores. Insights from this study advance understanding of EFL students' use of generative AI in expository writing. By revising expectations and strategies for integrating generative AI tools in expository writing pedagogy, educators can equip students to become more sophisticated expository writers and responsible generative AI users.

Vandermeulen, N., Lindgren, E., Waldmann, C., & Levlin, M. (2024). Getting a grip on the writing process:(Effective) approaches to write argumentative and narrative texts in L1 and L2. *Journal of Second Language Writing*, *65*, 101113. https://doi.org/10.1016/j.jslw.2024.101113

Wang, C. (2024). Exploring Students' Generative AI-Assisted Writing Processes: Perceptions and Experiences from Native and Nonnative English Speakers. *Technology, Knowledge and Learning*. https://doi.org/10.1007/s10758-024-09744-3

White, J., Fu, Q., Hays, S., Sandborn, M., Olea, C., Gilbert, H., Elnashar, A., Spencer-Smith, J., & Schmidt, D. C. (2023). *A Prompt Pattern Catalog to Enhance Prompt Engineering with ChatGPT* (arXiv:2302.11382). arXiv. http://arxiv.org/abs/2302.11382

White, M. D., & Marsh, E. E. (2006). Content Analysis: A Flexible Methodology. *Library Trends, 55*(1), 22–45. https://doi.org/10.1353/lib.2006.0053

Wolz, U., Stone, M., Pearson, K., Pulimood, S. M., & Switzer, M. (2011). Computational thinking and expository writing in the middle school. *ACM Transactions on Computing Education (TOCE)*, *11*(2), 1–22. https://doi.org/10.1145/1993069.1993073

Woo, D. J., Guo, K., & Susanto, H. (2024a). Exploring EFL students' prompt engineering in human-AI story writing: An activity theory perspective. *Interactive Learning Environments,* 1–20. https://doi.org/10.1080/10494820.2024.2361381

Woo, D. J., Susanto, H., Yeung, C. H., Guo, K., & Fung, A. K. Y. (2024b). Exploring AI-Generated text in student writing: How does AI help? *Language Learning & Technology*, *28*(2), 183–209. https://hdl.handle.net/10125/73577

Woo, D. J., Wang, D., Guo, K., & Susanto, H. (2024c). Teaching EFL students to write with ChatGPT: Students' motivation to learn, cognitive load, and satisfaction with the learning process. *Education and Information Technologies*. https://doi.org/10.1007/s10639-024-12819-4

Wu, Y., & Schunn, C. D. (2021). The effects of providing and receiving peer feedback on writing performance and learning of secondary school students. *American Educational Research Journal*, *58*(3), 492–526. https://doi.org/10.3102/0002831220945266

Yang, D., Zhou, Y., Zhang, Z., & Li, T. J.-J. (2022). AI as an active writer: Interaction strategies with generated text in human–AI collaborative fiction writing. *Joint Proceedings of the ACM IUI Workshops 2022*, 10.

**Appendix 1.** Final Coding Scheme.

| Set | Code | Description | Example |
|---|---|---|---|
| Expository Writing Unit | Title | The name of the composition. Delimited by a line break | Discover the Serenity and Health Benefits of Tai Chi: A Hidden Gem for Tourists |
| Expository Writing Unit | Heading | The name of a section (e.g. paragraph) in a composition. Delimited by a line break | The Tale of Mr Wu |
| Expository Writing Unit | Topic Sentence | The first sentence of a body paragraph | At the heart of our conversation, Mr Wu painted a vivid picture of the initial shockwave that reverberated through the industry. |
| Expository Writing Unit | Supporting Sentence | The second and subsequent sentences in a body paragraph | Forced closures, plummeting revenues, and the uncertainty of the future cast a dark cloud over their once-thriving establishment. |
| Expository Writing Unit | Introductory Paragraph | The first paragraph of a composition. Delimited by line breaks | 2020. The year that tested the world's resilience. Among the many sectors profoundly impacted by the global pandemic, the restaurant industry bore the brunt of unprecedented challenges. However, from the ruins rose a resilient exception, over which now a ray of dawn light shines after the merciless storm. In this exclusive interview with Mr Wu, the owner of Ho Ging Cha Chaan Teng, we had the privilege of taking in valuable information about his story, imbued with tenacity, valiance, and promise. |
| Expository Writing Unit | Concluding Paragraph | The last paragraph of a composition. Delimited by line breaks | It had been strenuous enough for these restaurant owners to run an entire restaurant, let alone during the pandemic, when the entire city seemed to shut down. Nonetheless, Mr Wu and his restaurant's story of persistence has attested to the significance of perseverance and innovation. Something we all should strive for. |
| Expository Writing Unit | Other | An unconventional or inappropriate expository writing unit (e.g. a letter greeting) | Dear esteemed readers, |
| Edit Scope | Short | Less than expository writing unit (e.g. words in a topic sentence); when a short chunk of a unit includes other units (e.g. short chunk of supporting sentence precedes short chunk of topic sentence of next paragraph), code initial unit's short chunk | Throughout the pandemic, |
| Edit Scope | Medium | Exactly one unit (e.g. a title); a sentence is delimited by a capital letter and a full stop, question mark or exclamation mark, whether or not the grammar within the sentence is accurate. Transforming a semantic unit (e.g. Heading) into another semantic unit (e.g. topic sentence) should be coded here. | Discover the Serenity and Health Benefits of Tai Chi: A Hidden Gem for Tourists |
| Edit Scope | Long | Exceeding one unit (e.g. a title and an introductory paragraph sentence) | Title: A Vibrant Celebration of Anime Culture<br><br>The Hong Kong Convention and Exhibition Centre was freshly transformed into a haven for anime enthusiasts as it hosted the illustrious Anime Expo. |

**Appendix 2.** HKDSE English Language Paper Two (Writing) Marking Scheme.

| Marks | Content (C) | Language (L) | Organization (O) |
|---|---|---|---|
| 7 | · Content entirely fulfills the requirements of the question<br>· Totally relevant<br>· All ideas are well developed/supported<br>· Creativity and imagination are shown when appropriate<br>· Shows a high awareness of audience | · Very wide range of accurate sentence structures, with a good grasp of more complex structures<br>· Grammar accurate with only very minor slips<br>· Vocabulary well-chosen and often used appropriately to express subtleties of meaning<br>· Spelling and punctuation are almost entirely correct<br>· Register, tone and style are entirely appropriate to the genre and text-type | · Text is organized extremely effectively, with logical development of ideas<br>· Cohesion in most parts of the text is very clear<br>· Cohesive ties throughout the text are sophisticated<br>· Overall structure is coherent, extremely sophisticated and entirely appropriate to the genre and text-type |
| 6 | · Content fulfills the requirements of the question<br>· Almost totally relevant<br>· Most ideas are well developed/supported<br>· Creativity and imagination are shown when appropriate<br>· Shows general awareness of audience | · Wide range of accurate sentence structures with a good grasp of simple and complex sentences<br>· Grammar mainly accurate with occasional common errors that do not affect overall clarity<br>· Vocabulary is wide, with many examples of more sophisticated lexis<br>· Spelling and punctuation are mostly correct<br>· Register, tone and style are appropriate to the genre and text-type | · Text is organized effectively, with logical development of ideas<br>· Cohesion in most parts of the text is clear<br>· Strong cohesive ties throughout the text<br>· Overall structure is coherent, sophisticated and appropriate to the genre and text-type |
| 5 | · Content addresses the requirements of the question adequately<br>· Mostly relevant<br>· Some ideas are well developed/supported<br>· Creativity and imagination are shown in most parts when appropriate<br>· Shows some awareness of audience | · A range of accurate sentence structures with some attempts to use more complex sentences<br>· Grammatical errors occur in more complex structures but overall clarity not affected<br>· Vocabulary is moderately wide and used appropriately<br>· Spelling and punctuation are sufficiently accurate to convey meaning<br>· Register, tone and style are mostly appropriate to the genre and text-type | · Text is mostly organized effectively, with logical development of ideas<br>· Cohesion in most parts of the text is clear<br>· Sound cohesive ties throughout the text<br>· Overall structure is coherent and appropriate to the genre and text-type |
| 4 | · Content just satisfies the requirements of the question<br>· Relevant ideas but may show some gaps or redundant information<br>· Some ideas but not well developed<br>· Some evidence of creativity and imagination | · Simple sentences are generally accurately constructed.<br>· Occasional attempts are made to use more complex sentences.<br>· Structures used tend to be repetitive in nature<br>· Grammatical errors sometimes affect meaning | · Parts of the text have clearly defined topics<br>· Cohesion in some parts of the text is clear<br>· Some cohesive ties in some parts of the text<br>· Overall structure is mostly coherent and appropriate to the genre and text-type |
| | · Shows occasional awareness of audience | · Common vocabulary is generally appropriate<br>· Most common words are spelt correctly, with basic punctuation being accurate<br>· There is some evidence of register, tone and style appropriate to the genre and text-type | |
| 3 | · Content partially satisfies the requirements of the question<br>· Some relevant ideas but there are gaps in candidates' understanding of the topic<br>· Ideas not developed, with possible repetition<br>· Does not orient reader effectively to the topic | · Short simple sentences are generally accurate.<br>· Only scattered attempts at longer, more complex sentences<br>· Grammatical errors often affect meaning<br>· Simple vocabulary is appropriate<br>· Spelling of common words is correct, with basic punctuation mostly accurate | · Parts of the text are generally defined<br>· Some simple cohesive ties used in some parts of the text but cohesion is sometimes fuzzy<br>· A limited range of cohesive devices are used appropriately |
| 2 | · Content shows very limited attempts to fulfil the requirements of the question<br>· Intermittently relevant<br>· Some ideas but few are developed<br>· Ideas may include misconception of the task or some inaccurate information<br>· Very limited awareness of audience | · Some short simple sentences accurately structured<br>· Grammatical errors frequently obscure meaning<br>· Very simple vocabulary of limited range often based on the prompt(s)<br>· A few words are spelt correctly with basic punctuation being occasionally accurate | · Parts of the text reflect some attempts to organize topics<br>· Some use of cohesive devices to link ideas |
| 1 | · Content inadequate and heavily based on the task prompt(s)<br>· A few ideas but none developed<br>· Some points/ ideas are copied from the task prompt or the reading texts<br>· Almost total lack of awareness of audience | · Multiple errors in sentence structures, spelling and/or word usage, which make understanding impossible | · Some attempt to organize the text<br>· Very limited use of cohesive devices to link ideas |
| 0 | · Totally inadequate<br>· Totally irrelevant or memorized<br>· All ideas are copied from the task prompt or the reading texts<br>· No awareness of audience | · Not enough language to assess | · Mainly disconnected words, short note-like phrases or incomplete sentences<br>· Cohesive devices almost entirely absent |

**Appendix 3a**. Final MLR Models Predicting Student Composition Qualities (Process-oriented, Eight-variable, Backward Elimination).

| Predictor | Content score model | | | | Language score model | | | | Organization score model | | | | Total score model | | | |
|---|---|---|---|---|---|---|---|---|---|---|---|---|---|---|---|---|
| | Standardized β | t | p | VIF | Standardized β | t | p | VIF | Standardized β | t | p | VIF | Standardized β | t | p | VIF |
| No. of Title edits | - | - | - | - | 0.244 | 1.230 | 0.232 | 1.200 | - | - | - | - | - | - | - | - |
| No. of Heading edits | 0.745 | 2.15 | 0.043 | 3.222 | 0.814 | 2.440 | 0.024 | 3.391 | - | - | - | - | 0.8 | 2.344 | 0.029 | 3.222 |
| No. of ConcludingP. edits | -0.715 | -2.066 | 0.051 | 3.222 | -0.84 | -2.583 | 0.017 | 3.223 | - | - | - | - | -0.739 | -2.164 | 0.042 | 3.222 |
| No. of Supporting S edits | - | - | - | - | - | - | - | - | -0.308 | -1.552 | 0.134 | 1 | - | - | - | - |

*Note.* Content score model: F(2, 22) = 2.440, *p* = .110; R² = .182; Adjusted R² = .107. The final model includes 2 predictors. Six predictors initially entered were excluded during the backward elimination procedure.

Language score model: F(3,21) = 3.153, p = .046; R² = .311; Adjusted R² = .212. The final model includes 3 predictors. Five predictors initially entered were excluded during the backward elimination procedure.

Organization score model: F(1,23) = 2.407, p = .134 ; R² = .095; Adjusted R² = .055. The final model includes 1 predictor. Seven predictors initially entered were excluded during the backward elimination procedure.

Total score model: F(2, 22) = 2.823, p = .081; R²= .204; Adjusted R² = .132. The final model includes 2 predictors. Six predictors initially entered were excluded during the backward elimination procedure.

**Appendix 3b.** Final MLR Models Predicting Student Composition Qualities (Process-oriented, Five-variable, Backward Elimination).

| | Content score model | | | | Language score model | | | | Organization score model | | | | Total score model | | | |
|---|---|---|---|---|---|---|---|---|---|---|---|---|---|---|---|---|
| Predictor | Standardized β | t | *p* | VIF | Standardized β | t | *p* | VIF | Standardized β | t | *p* | VIF | Standardized β | t | *p* | VIF |
| No. of Supporting S (Short) edits | -0.193 | -0.941 | 0.356 | 1 | -0.164 | -0.796 | 0.434 | 1 | -0.249 | -1.234 | 0.23 | 1 | -0.209 | -1.206 | 0.316 | 1 |

\* *Note.* Content score model: F(1, 23) = .886 , p = .356; R² = .037; Adjusted R² = -.005. The final model includes 1 predictor. Four predictors initially entered were excluded during the backward elimination procedure.

Language score model: F(1, 23) = .634 , p = .434; R² = .027; Adjusted R² = -.015. The final model includes 1 predictor. Four predictors initially entered were excluded during the backward elimination procedure.

Organization score model: F(1, 23) = 1.524, p = .230; R² = .062; Adjusted R² = .021. The final model includes 1 predictor. Four predictors initially entered were excluded during the backward elimination procedure.

Total score model: F(1, 23) = 1.053 p = .0.316; R² = .044; Adjusted R² = .002.The final model includes 1 predictor. Four predictors initially entered were excluded during the backward elimination procedure.

**Appendix 3c.** Summary of Partial Correlation in the Five-variable Analyses (Process-oriented).

|  | Content score | Language score | Organization score | Total score |
|---|---|---|---|---|
| **Total runtime** | - | - | - | - |
| **No. of Title (Medium) edits** | - | - | - | - |
| **No. of Topic S (Long) edits** | - | - | - | - |
| **No. of Supporting S (Short) edits** | -0.193 | -0.164 | -0.249 | -0.209 |
| **No. of Intro P (short) edits** | - | - | - | - |
| **Model *p*-value** | 0.356 | 0.434 | 0.230 | 0.316 |
| ***R*-squared value** | 0.037 | 0.027 | 0.062 | 0.044 |
| **Adjusted *R*-squared value** | - 0.005 | - 0.015 | 0.021 | 0.002 |

**Appendix 4a.** Final MLR Models Predicting Student Composition Qualities (Product Perspective, Nine-variable, Enter).

| Predictor | Content score model | | | | Language score model | | | | Organization score model | | | | Total score model | | | |
|---|---|---|---|---|---|---|---|---|---|---|---|---|---|---|---|---|
| | Standardized β | t | *p* | VIF | Standardized β | t | *p* | VIF | Standardized β | t | *p* | VIF | Standardized β | t | *p* | VIF |
| No. of Human words | 0.375 | 1.892 | 0.069 | 2.466 | 0.371 | 1.83 | 0.078 | 2.466 | 0.418 | 1.956 | 0.06 | 2.466 | 0.399 | 1.999 | 0.055 | 2.466 |
| No. of AI words | 0.419 | 2.473 | 0.019 | 1.806 | 0.48 | 2.765 | 0.01 | 1.806 | 0.399 | 2.181 | 0.037 | 1.806 | 0.445 | 2.607 | 0.014 | 1.806 |
| No. of Title chunks | -0.127 | -0.807 | 0.426 | 1.552 | -0.151 | -0.94 | 0.355 | 1.552 | -0.109 | -0.644 | 0.525 | 1.552 | -0.133 | -0.84 | 0.408 | 1.552 |
| No. of Heading chunks | -0.446 | -3.284 | 0.003 | 1.157 | -0.454 | -3.271 | 0.003 | 1.157 | -0.451 | -3.083 | 0.004 | 1.157 | -0.464 | -3.394 | 0.002 | 1.157 |
| No. of Topic S. chunks | 0.282 | 1.634 | 0.113 | 1.865 | 0.198 | 1.122 | 0.271 | 1.865 | 0.161 | 0.865 | 0.394 | 1.865 | 0.225 | 1.295 | 0.205 | 1.865 |
| No. of Supporting S. chunks | -0.107 | -0.336 | 0.74 | 6.345 | -0.076 | -0.233 | 0.817 | 6.345 | -0.103 | -0.302 | 0.765 | 6.345 | -0.099 | -0.309 | 0.759 | 6.345 |
| No. of IntroP. chunks | -0.005 | -0.035 | 0.973 | 1.454 | -0.085 | -0.547 | 0.588 | 1.454 | -0.103 | -0.628 | 0.535 | 1.454 | -0.062 | -0.407 | 0.687 | 1.454 |
| No. of ConcludingP. chunks | 0.139 | 0.528 | 0.602 | 4.371 | 0.108 | 0.401 | 0.692 | 4.371 | 0.118 | 0.415 | 0.681 | 4.371 | 0.127 | 0.477 | 0.637 | 4.371 |
| No. of Other chunks | -0.256 | -1.999 | 0.055 | 1.03 | -0.264 | -2.01 | 0.054 | 1.03 | -0.319 | -2.312 | 0.028 | 1.03 | -0.286 | -2.221 | 0.034 | 1.03 |

*Note.* Content score model: $F_{(9,29)} = 3.756$, $p = .003$; $R^2 = .538$; Adjusted $R^2 = .395$;
Language score model: $F_{(9,29)} = 3.440$, $p = .005$; $R^2 = .516$; Adjusted $R^2 = .366$;
Organization score model: $F_{(9,29)} = 2.783$, $p = .018$; $R^2 = .463$; Adjusted $R^2 = .297$;
Total score model: $F_{(9,29)} = 3.662$, $p = .004$; $R^2 = .532$; Adjusted $R^2 = .387$.

**Appendix 4b.** Final MLR Models Predicting Student Composition Quality - Content Score (Product Perspective, six-variable, Backward Elimination).

| Predictor | Content score model | | | | Language score model | | | | Organization score model | | | | Total score model | | | |
|---|---|---|---|---|---|---|---|---|---|---|---|---|---|---|---|---|
| | Standardized β | t | p | VIF | Standardized β | t | p | VIF | Standardized β | t | p | VIF | Standardized β | t | p | VIF |
| No. of Human words | 0.335 | 1.797 | 0.081 | 1.565 | 0.317 | 1.684 | 0.101 | 1.565 | 0.363 | 1.872 | 0.069 | 1.565 | 0.343 | 1.811 | 0.079 | 1.565 |
| No. of AI words | 0.378 | 1.979 | 0.056 | 1.643 | 0.448 | 2.321 | 0.026 | 1.643 | 0.417 | 2.154 | 0.038 | 1.643 | 0.413 | 2.126 | 0.041 | 1.643 |
| No. of Topic S (short) chunks | 0.282 | 1.835 | 0.075 | 1.065 | 0.199 | 1.28 | 0.209 | 1.065 | - | - | - | - | 0.214 | 1.368 | 0.18 | 1.065 |

*Note.* Content score model: $F(3,35) = 3.354$, $p = .03$; $R^2 = .223$; Adjusted $R^2 = .157$. The final model includes 3 predictors. Three predictors initially entered were excluded during the backward elimination procedure.

Language score model: $F(3,35) = 3.030$, $p = .042$; $R^2 = .206$; Adjusted $R^2 = .138$. The final model includes 3 predictors. Three predictors initially entered were excluded during the backward elimination procedure.

Organization score model: $F(2,36) = 2.591$, $p = .089$; $R^2 = .126$; Adjusted $R^2 = .077$. The final model includes 2 predictors. Four predictors initially entered were excluded during the backward elimination procedure.

Total score model: $F(3,35) = 2.882$, $p = .05$; $R^2 = .198$; Adjusted $R^2 = .129$. The final model includes 3 predictors. Three predictors initially entered were excluded during the backward elimination procedure.

**Appendix 4c**. Summary of Partial Correlation in the Six-variable Analyses (Product-oriented).

|  | Content score | Language score | Organization score | Total score |
|---|---|---|---|---|
| **No. AI words** | 0.317† | 0.365* | 0.338* | 0.338* |
| **No. Human words** | 0.291† | 0.274 | 0.298† | 0.293† |
| **No. Topic S(Short)** | 0.296† | 0.211 | - | 0.225 |
| **No. Supporting S(Short)** | - | - | - | - |
| **No. IntroP(Short)** | - | - | - | - |
| **No. ConcludingP(Short)** | - | - | - | - |
| **Model *p*-value** | **0.03** | **0.042** | 0.089 | **0.05** |
| ***R*-squared value** | 0.223 | 0.206 | 0.126 | 0.198 |
| **Adjusted *R*-squared value** | 0.157 | 0.138 | 0.077 | 0.129 |

*Note*: "*" means the partial correlation value is significant with $p < .05$, and "†" means $.05 < p < .10$, marginally significant.